\documentclass{article}
\usepackage{ijcai24}
\usepackage{balance}
\usepackage{times}
\usepackage{soul}
\usepackage{url}
\usepackage[hidelinks]{hyperref}
\usepackage[utf8]{inputenc}
\usepackage[small]{caption}
\usepackage{graphicx}
\usepackage{amsmath}
\usepackage{amsthm}
\usepackage{booktabs}
\usepackage{algorithm}
\usepackage{algorithmic}
\usepackage[switch]{lineno}
\usepackage{float}
\usepackage{bm}
\usepackage{amssymb}
\usepackage{color}
\usepackage{todonotes}
\usepackage{multirow} 
\usepackage{multicol} 
\usepackage{booktabs}
\usepackage{color}
\usepackage{tabularray}
\usepackage{tabularx} 
\usepackage{hhline}
\usepackage{textcomp}
\usepackage{placeins}
\usepackage{amssymb}
\usepackage{tabularx}

\title{Hierarchical Space-Time Attention for Micro-Expression Recognition}

\author{
Haihong Hao$^1$
\and
Shuo Wang$^{1}$\footnote{Corresponding author}\and
Huixia Ben$^2$\and
Yanbin Hao$^1$\and
Yansong Wang $^3$\And
Weiwei Wang$^3$
\affiliations
$^1$University of Science and Technology of China\\
$^2$Hefei University of Technology\\
$^3$Chery HuiYin Motor Finance Service Co.,Ltd\\
\emails
haohaihong@mail.ustc.edu.cn,
\{shuowang.edu,huixiaben\}@gmail.com,
haoyanbin@hotmail.com,
\{wangyansong,wangweiwei\}@cheryfs.cn
}
\begin{document}

\maketitle
\begin{abstract}
Micro-expression recognition (MER) aims to recognize the short and subtle facial movements from the Micro-expression (ME) video clips, which reveal real emotions. Recent MER methods mostly only utilize special frames from ME video clips or extract optical flow from these special frames. However, they neglect the relationship between movements and space-time, while facial cues are hidden within these relationships.
To solve this issue, we propose the \textbf{H}ierarchical \textbf{S}pace-\textbf{T}ime \textbf{A}ttention (HSTA). Specifically, we first process ME video frames and special frames or data parallelly by our cascaded \textbf{U}nimodal \textbf{S}pace-\textbf{T}ime \textbf{A}ttention (USTA) to establish connections between subtle facial movements and specific facial areas. Then, we design \textbf{C}rossmodal \textbf{S}pace-\textbf{T}ime \textbf{A}ttention (CSTA) to achieve a higher-quality fusion for crossmodal data. Finally, we hierarchically integrate USTA and CSTA to grasp the deeper facial cues. Our model emphasizes temporal modeling without neglecting the processing of special data, and it fuses the contents in different modalities while maintaining their respective uniqueness. Extensive experiments on the four benchmarks show the effectiveness of our proposed HSTA. Specifically, compared with the latest method on the CASME3 dataset, it achieves about 3\% score improvement in seven-category classification. Code is available at \url{https://github.com/OceanSummerDay/HSTA_MER}.

\end{abstract}

\section{Introduction}


Micro-expression recognition (MER) is a challenging task in affective computing, due to the subtlety and brief duration (typically 1/25 to 1/3 second) of micro-expressions (MEs), making them difficult to capture \cite{van2019capsulenet}. However, MEs are brief, subtle, spontaneous, and involuntary emotional expressions that convey genuine emotions. Although humans can hide their emotions in certain situations, MEs are difficult to hide and can inadvertently  reveal true feelings \cite{ekman2009telling}. Thus, the MER is critically important in many specific scenes, such as criminal interrogation, clinical diagnosis, and financial risk control. 


Recent advances in deep learning have shown promising results in MER \cite{verma2019learnet}, surpassing traditional hand-crafted methods and emerging as the dominant technique. However, the majority of open-source methods overly depend on special frames or specific data, failing to fully utilize the fundamental temporal characteristics of MEs as special short video sequences. Techniques like CapsuleNet \cite{van2019capsulenet}, MMNet \cite{li2022mmnet}, and OFF-ApexNet \cite{gan2019off} rely on special frames (Apex and Onset frame),  neglecting the most intrinsic video nature of ME data. Similarly, methods such as Dual-ATME \cite{liong2017micro}, DSSN \cite{khor2019dual}, and Bi-WOOF \cite{liong2018less} rely heavily on special optical flow, which is highly sensitive to environmental factors and is unable to fully capture all the details of non-rigid and complex facial muscle movements. These methods commonly lack the ability to model temporal information effectively, leading to a disconnection between facial movements and specific facial areas. The model only associate facial expressions with static facial features, rather than interpreting ME as a series of continuous facial motions. Actually, MER requires a more comprehensive modeling of space-time information to better capture the correspondence between dynamic facial muscle movements and specific facial regions, which is essential for accurately recognizing micro-expressions \cite{li2022mmnet}. 


For temporal (space-time) modeling, we note that existing video processing methods \cite{wang2023videomae,rehman2022federated,qian2021spatiotemporal} have achieved achieved great success on conventional action recognition datasets. For both action recognition and MER tasks, it is essential to establish the relationship between movements and time, as well as to identify the connections between specific actions and certain areas or scenes. 
Inspired by these action recognition work, we incorporate these temporal processing concepts into the MER task. 
Specifically, we extract both spatial and temporal information by our designed \textbf{U}nimodal \textbf{S}pace-\textbf{T}ime \textbf{A}ttention (USTA) from a small number of uniformly sampled micro-expression frames. This is because the key is to establish connections between these tiny muscle movements and specific small areas of face \cite{li2022mmnet}. By incorporating and modeling temporal information, USTA can capture tiny movements over time since all tokens (features) are able to interact with each other \cite{vaswani2017attention} in our USTA. In other words, tokens of the same facial area can interact with their counterparts at different moments and capture subtle facial movements over time. Meanwhile, tokens for different facial areas can also interact with each other and associate minute movements with specific areas of the face.


Based on the studies from past MER methods, we also recognize that the special frames or data are important for recognizing the MEs. Thus, we design the \textbf{C}rossmodal \textbf{S}pace-\textbf{T}ime \textbf{A}ttention (CSTA) to effectively fuse the information from different modalities after the unimodal USTA calculations. Specifically, we utilize a symmetrical structure based on cross-attention to capture the inner connection between ME video frames and special data (such as special frames or optical flow). In CSTA, two different modalities of data (temporal video data and special data) are captured and integrated by our cross-attention calculations. Meanwhile, these data also complement each other to assist in the expression of emotions.
After passing through our CSTA, the class tokens contain information from the other type of data, while the remaining tokens retain their original data. Thus, the combination of USTA and CSTA achieves effective fusion while maintaining the distinctiveness of different modalities. To adequately grasp the deeper facial cues of motion and time, we extend this combination into a hierarchical structure, named \textbf{H}ierarchical \textbf{S}pace-\textbf{T}ime \textbf{A}ttention (HSTA). In HSTA, the USTA and CSTA are stacked in an orderly manner to capture the MEs effectively. Meanwhile, the adaptable layer design of HSTA enhances its generalization capabilities. Our contributions can be summarized as threefold:
\begin{itemize}
\item We design unimodal space-time attention (USTA) to demonstrate the significance of temporal information in MER. Meanwhile, we propose crossmodal space-time attention (CSTA) to complement data of different types, thereby enriching the content within each modality data.

\item We extend the USTA and CSTA into the hierarchical structure to fuse the contents in different modalities and grasp the deeper facial cues of motion and temporal data.

\item We demonstrate the effectiveness of our proposed HSTA on four MER datasets. Our method outperforms existing methods and achieves state-of-the-art (SOTA) results.
\end{itemize}

\section{Related Work}
In this section, we first briefly introduce common solutions for MER tasks and then we list the related attention calculations. Finally, we enumerate the differences between our methods and those of related methods. 

\subsection{Micro-expression Recognition}
MER task is to classify the facial MEs in a video. Related technologies are mainly divided into two categories. The \textbf{first category} relies solely on special data (e.g., special frames, optical flow) and feeds them into a 2D CNN. This approach is the one most commonly adopted in most current open-source work. They are highly sensitive to environmental factors \cite{zhou2019dual} such as changes in lighting, shadows. In the \textbf{other category}, temporal MEs are input into the model and then learned by a time series network or a 3D CNN (\cite{reddy2019spontaneous,khor2018enriched}). However, due to information redundancy, it becomes difficult to focus on the most important features. Consequently, the performance of these methods is probably not as effective as that of those using only special data, leading to their relative underestimation. Although there have been attempts to model temporal sequences of optical flow \cite{li2019micro}, due to the subtlety of MEs and minimal changes between adjacent frames, extracting optical flow results in significant noise, leading to poor performance. In reality, we can model temporal sequences while minimizing redundancy, without neglecting the importance of special data.

Modeling spatial and temporal information is key to processing sequential data. Currently, the main methods include 3D CNNs, Video Vision Transformers \cite{arnab2021vivit} (ViTs), or a combination of both \cite{wodajo2021deepfake}. A 3D CNN extends the standard convolution operation from 2D to 3D, allowing it to capture motion information embedded within consecutive frames of a video by analyzing a sequence as a whole, rather than in isolation. ViTs capture global dependencies in video frames, providing a broader understanding of the scene compared to the local focus of 3D CNNs. With their self-attention mechanism, ViTs dynamically concentrate on relevant parts of the input, which is significant for processing inputs from various modalities. However, a direct application of these video models presents drawbacks: they only model temporal sequences of videos, neglecting the importance of special frames. Compared to methods utilizing optical flow, these methods also lack the capture of the big picture. 

\subsection{Attention Calculation}
 As a method similar to \cite{tong2022videomae}, employing self-attention mechanisms on temporal data enables the capture of subtle movements and their temporal relationships. As methods such as \cite{khor2019dual} have shown, simply adding or concatenating feature vectors from different modalities does not significantly improve performance, so we should find a better way to fuse different modalities. Cross-attention has already been applied to non-sequential data \cite{wei2020multi}, exploiting crossmodal relationships and achieving tremendous success. Cross-attention between two different modalities offers a higher quality fusion, integrating the different modalities more effectively.

Based on the analysis of related work, the methods most related to ours are the recently proposed cross-attention \cite{chen2021crossvit} and dual learning in different modalities \cite{khor2019dual}. Our method differs from theirs in two aspects. First, we design a \textbf{new attention strategy} for different temporal MEs data and process them parallel, to establish the relationship between motions and time. Second, we introduce a \textbf{new cross-attention} to achieve high-quality fusion for sequential data, still keeping their uniqueness to model connections between different modalities.

\begin{figure*}[t]
\centering
\includegraphics[width=0.98\textwidth]{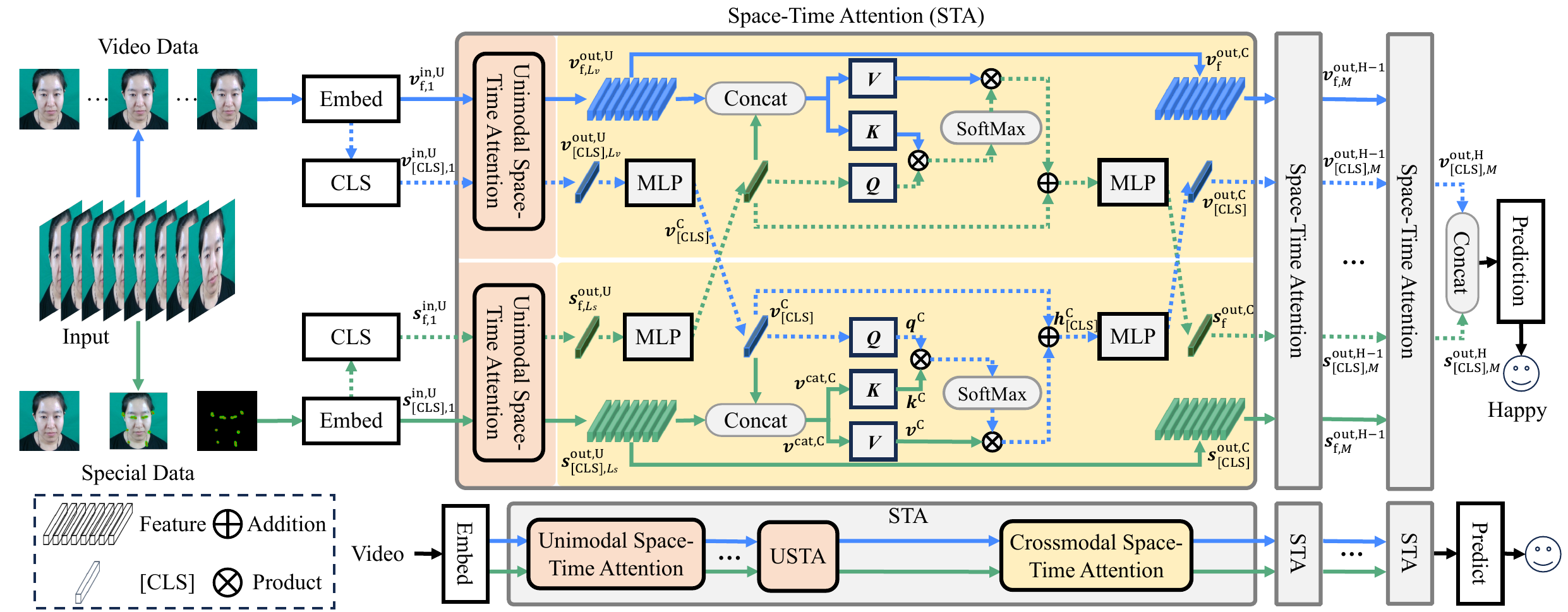}
\caption{The overview of our Hierarchical Space-Time Attention (HSTA). Our model is a hierarchical structure that captures the underlying emotions in micro-expression videos through multiple attention modules.}
\label{fig:pipeline}
\end{figure*}

\section{Approach}
In this section, we first briefly revisit the preliminaries of the MER tasks and give an overview of our framework. Then, we illustrate our Hierarchical Space-Time Attention (HSTA) containing Unimodal Space-Time Attention (USTA) and Crossmodal Space-Time Attention (CSTA). Meanwhile, we introduce the design of hierarchical structures. Finally, we describe the training and inference procedures of our method.



\subsection{Preliminaries}


The data of the MER task can be divided into two parts: \textbf{video frames set} $\mathcal{D}_{v}$ and \textbf{special frames set} $\mathcal{D}_{s}$. Specifically, $\mathcal{D}_{v}$ consists of multiple video clips, and each clip provides the label of its emotional expression. $ \mathcal{D}_{s}$ contain some special frames, such as Apex, Onset, and Offset frames, where the Apex, Onset, and Offset frames represent the moments of maximum expression amplitude, the start, and the end of a micro-expression video. They can be used to assist in the recognition of facial expression content. The goal of the MER task is to use these different formats of data to capture the emotional content. 

An overview of our method is depicted in Figure \ref{fig:pipeline}. First, we use available embed methods to capture the features and the ``[CLS]s'' of video and special frames, where ``[CLS]'' can be regarded as features containing certain classification information. Then we design the unimodal and crossmodal space-time attentions to fuse these different contents. Meanwhile, we expand these attentions into hierarchical structures to obtain a deep emotional expression. Finally, we concatenate the ``[CLS]s'' from different modalities for prediction.

\subsection{Unimodal Space-Time Attention }
In our method, video and special frames are calculated simultaneously. Since the Unimodal Space-Time Attention (USTA) calculations of these two parts are similar, we first introduce the operation of USTA and then describe its application in these two parts' calculations to simplify the description.

\begin{figure}[t]
    \centering
    \includegraphics[width=0.48\textwidth]{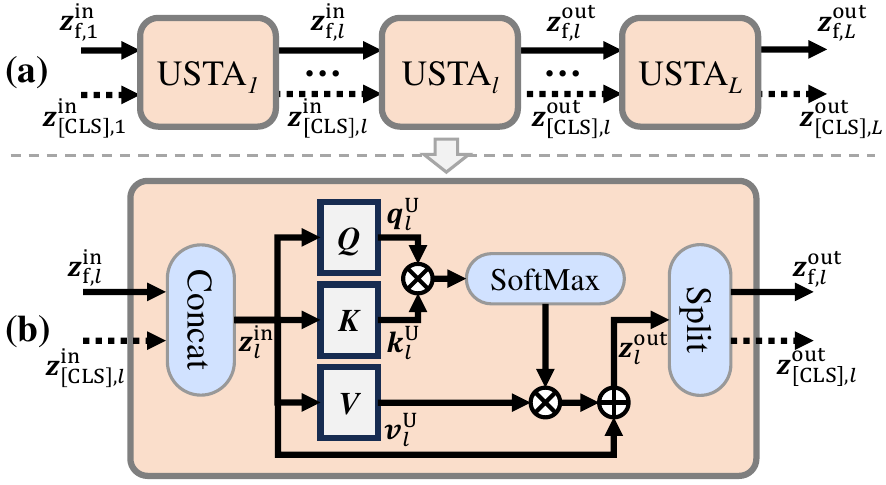}
    \caption{The details of USTA. (a) The cascaded structure of USTA. (b) The calculations of $l^{\rm th}$ layer USTA.}
    \label{fig:USTA}
\end{figure}

As shown in Figure~\ref{fig:USTA}(a), our USTA is a \textbf{cascaded} structure. To illustrate it in detail, we use the $l^{\rm th}$ layer of calculations as an example to describe each USTA unit. Details are depicted in Figure~\ref{fig:USTA}(b), denoting the inputted features and ``[CLS]'' of the $l^{\rm th}$ layer as $\bm{z}_{{\rm f},l}^{\rm in} \in\mathbb{R}^{N\times d}$ and $\bm{z}_{{\rm [CLS]},l}^{\rm in} \in\mathbb{R}^{d}$, respectively,
where $N$ is the number of features and $d$ is the size of the embedding, we first concatenate them into one feature:
\begin{equation}
\bm{z}_{l}^{\rm in} = [\bm{z}_{{\rm f},l}^{\rm in} || \bm{z}_{{\rm [CLS]},l}^{\rm in}] \in \mathbb{R}^{(N + 1)\times d}.
\end{equation}
Then we design three attention matrices $\bm{q}^{\rm U}_{l}$, $\bm{k}^{\rm U}_{l}$, and $\bm{v}^{\rm U}_{l}$ to capture self-similarities from the concatenated feature $\bm{z}_{l}^{\rm in}$:
\begin{equation}
\bm{q}^{\rm U}_{l}= \bm{z}_{l}^{\rm in} \bm{W}_{q,l}^{\rm U}, \ 
\bm{k}^{\rm U}_{l}= \bm{z}_{l}^{\rm in} \bm{W}_{k,l}^{\rm U}, \ 
\bm{v}^{\rm U}_{l}= \bm{z}_{l}^{\rm in} \bm{W}_{v,l}^{\rm U}, 
\end{equation}
where $\bm{W}_{q,l}^{\rm U}, \bm{W}_{k,l}^{\rm U}$, and $\bm{W}_{v,l}^{\rm U}$ are learnable parameters in $ \mathbb{R}^{d\times d}$, and $\bm{q}^{\rm U}_{l}, \bm{k}^{\rm U}_{l}$, and $\bm{v}^{\rm U}_{l}$ are same size to the input $\bm{z}_{l}^{\rm in}$. Third, following the operations in \cite{vaswani2017attention}, we employ self-attention to achieve the refinement feature $\bm{z}_{l}^{\rm out}$:
\begin{equation}
\bm{z}^{\rm out}_{l} = {\rm LayerNorm}(\bm{z}_{l}^{\rm in} + {\rm SoftMax}(\frac{\bm{q}^{\rm U}_{l}(\bm{k}_{l}^{\rm U})^\intercal}{\sqrt{d}})\bm{v}^{\rm U}_{l}),
\end{equation}
where ``{\rm LayerNorm}'' and ``{\rm SoftMax}'' are normalization function and activation function. Finally, we split the refined feature $\bm{z}^{\rm out}_{l}\in\mathbb{R}^{(N + 1)\times d}$ into two parts as the outputs of the $l^{\rm th}$ layer USTA:
\begin{equation}
[\bm{z}_{{\rm f},l}^{\rm out} || \bm{z}_{{\rm [CLS]},l}^{\rm out}] = {\rm Split}(\bm{z}^{\rm out}_{l}), 
\end{equation}
where $\bm{z}_{{\rm f},l}^{\rm out}\in \mathbb{R}^{N\times d}$ and $\bm{z}_{{\rm [CLS]},l}^{\rm out}\in \mathbb{R}^{d}$ are the features and ``[CLS]'', respectively. During this operation, the contents of the frames along the time dimension can interact with each other. To conveniently describe our cascaded USTA, we define the calculation of USTA in the $l^{\rm th}$ layer as ${\rm USTA}_{l}$. The different layer calculations can be defined as:
\begin{equation}
[\bm{z}_{{\rm f},l}^{\rm out} || \bm{z}_{{\rm [CLS]},l}^{\rm out}] ={\rm USTA}_{l}([\bm{z}_{{\rm f},l}^{\rm in} || \bm{z}_{{\rm [CLS]},l}^{\rm in}]).
\end{equation}
For the cascade structure, we calculate the input and output of different USTA layers as $\bm{z}^{\rm out}_{L} = [\bm{z}_{{\rm f}, L}^{\rm out} || \bm{z}_{{\rm [CLS]}, L}^{\rm out}]$ :
\begin{equation}
\bm{z}_{L}^{\rm out} =
\begin{cases} 
[\bm{z}_{{\rm f},1}^{\rm in} || \bm{z}_{{\rm [CLS]},1}^{\rm in}], &L = 1, \\
{\rm USTA}_{L}([\bm{z}_{{\rm f},{L-1}}^{\rm out} || \bm{z}_{{\rm [CLS]},{L-1}}^{\rm out}]), &L > 1, 
\end{cases}
\label{eq:usta}
\end{equation}
where $L$ is the total layer of the cascaded USTA. Then we illustrate the USTA calculations in different modalities.

For given video frames set $\mathcal{D}_{v}$ and special frames set $\mathcal{D}_{s}$, we use different 3D CNN \cite{tong2022videomae} for feature extraction. With setting ``[CLS]'' token, the embedding of $\mathcal{D}_{v}$ and $\mathcal{D}_{s}$ can be represented as $[\bm{v}_{{\rm f},1}^{{\rm in,U}} || \bm{v}_{{\rm [CLS]},1}^{{\rm in,U}}] \in \mathbb{R}^{(N_v+1)\times d} $ and $[\bm{s}_{{\rm f},1}^{{\rm in,U}} || \bm{s}_{{\rm [CLS]},1}^{{\rm in,U}}] \in \mathbb{R}^{(N_s+1)\times d}$, where $N_v$ and $N_s$ are the number of features of $\mathcal{D}_{v}$ and $\mathcal{D}_{s}$ respectively. Then we define $L_v$ and $L_s$ as the total layer of USTA in these two calculations in Eq.~\eqref{eq:usta}. We process them \textbf{parallel}. Thus, the USTA outputs from $\mathcal{D}_{v}$ and $\mathcal{D}_{s}$ can be calculated as:
\begin{equation}
\begin{aligned}
[\bm{v}_{{\rm f},L_v}^{{\rm out,U}} || \bm{v}_{{\rm [CLS]},L_v}^{{\rm out,U}}] &= {\rm USTA}_{L_v}([\bm{v}_{{\rm f},1}^{{\rm in,U}} || \bm{v}_{{\rm [CLS]},1}^{{\rm in,U}}]), \\
[\bm{s}_{{\rm f},L_s}^{{\rm out,U}} || \bm{s}_{{\rm [CLS]},L_s}^{{\rm out,U}}] &= {\rm USTA}_{L_s}([\bm{s}_{{\rm f},1}^{{\rm in,U}} || \bm{s}_{{\rm [CLS]},1}^{{\rm in,U}}]).
\label{eq:lvls}
\end{aligned}
\end{equation}
In this operation, we use the same embedding size $d$ to facilitate our subsequent calculations. The outputs of USTA modules ($[\bm{v}_{{\rm f}, L_v}^{{\rm out,U}} || \bm{v}_{{\rm [CLS]}, L_v}^{{\rm out,U}}]$ and $[\bm{s}_{{\rm f}, L_s}^{{\rm out,U}} || \bm{s}_{{\rm [CLS]}, L_s}^{{\rm out,U}}]$) are then used as the input for crossmodal space-time attention.

\subsection{Crossmodal Space-Time Attention}



Crossmodal Space-Time Attention (CSTA) is designed to capture the inner connection between the features of one modal data and the ``[CLS]'' of another. Since the calculations in our CSTA from different modalities are \textbf{symmetrical}, we only discuss the calculation between the features of {special} data $\bm{s}_{{\rm f},L_s}^{\rm out,U}$ and ``[CLS]'' of {video} data $\bm{v}_{{\rm [CLS]},L_v}^{\rm out,U}$ for brevity description (in the \textbf{lower} part of the golden area of Figure \ref{fig:pipeline}, and the other part of the operations are similar). 

Firstly, we project ``[CLS]'' of video data by a MLP to match the embedding dimension of the features of special data and then concatenate them for the subsequent operation:
\begin{equation}
\bm{v}_{{\rm [CLS]}}^{{\rm C}} = {\rm MLP}_1(\bm{v}_{{\rm [CLS]},L_v}^{{\rm out,U}}), 
\bm{v}^{{\rm cat,C}} =[\bm{v}_{{\rm [CLS]}}^{{\rm C}} || \bm{s}_{{\rm f},L_s}^{{\rm out,U}}].
\end{equation}
Similarly to the calculation in the attention of the USTA, we then also design three attention matrices $\bm{q}^{\rm C}$, $\bm{k}^{\rm C}$, and $\bm{v}^{\rm C}$ to capture cross-similarities between ``[CLS]'' of video data $\bm{v}_{{\rm [CLS]}}^{{\rm C}}$ and the concatenated feature $\bm{v}^{{\rm cat, C}}$:
%
%
%
\begin{equation}
\bm{q}^{\rm C}= \bm{v}_{{\rm [CLS]}}^{{\rm C}} \bm{W}_{q}^{\rm C}, 
\bm{k}^{\rm C}= \bm{v}^{{\rm cat,C}} \bm{W}_{k}^{\rm C}, 
\bm{v}^{\rm C}= \bm{v}^{{\rm cat,C}} \bm{W}_{v}^{\rm C},
\label{eq:qkv}
\end{equation}
where the definition of $\bm{q}^{\rm C}$, $\bm{k}^{\rm C}$, $\bm{v}^{\rm C}$, $\bm{W}_{q}^{\rm C}, \bm{W}_{k}^{\rm C}$, and $\bm{W}_{v}^{\rm C}$ are similar to that of the USTA. Thus, the refinement ``[CLS]'' can be calculated by the attention and residual operations:
\begin{equation}
\bm{h}_{\rm [CLS]}^{\rm C} = \bm{v}_{{\rm [CLS]}}^{{\rm C}} + {\rm SoftMax}(\frac{\bm{q}^{\rm C}(\bm{k}^{\rm C})^\intercal}{\sqrt{d}}) \bm{v}^{\rm C}.
\end{equation}
Finally, we map $\bm{h}^{C}_{\rm [CLS]}$ by using a new MLP to keep the output dimensions consistent:
\begin{equation}
\bm{v}_{\rm [CLS]}^{\rm out,C} = {\rm MLP}_2(\bm{h}_{\rm [CLS]}^{\rm C}).
\end{equation}
Meanwhile, the output features of special data is consistent with the original inputted features:

\begin{equation}
\bm{s}_{\rm f}^{\rm out,C} = \bm{s}_{{\rm f},L_s}^{{\rm out,U}}.
\end{equation}

Another part has similar operations. Thus, given the features of {video} data $\bm{v}_{{\rm f},L_v}^{\rm out,U}$ and ``[CLS]'' of {special} data $\bm{s}_{{\rm [CLS]},L_s}^{\rm out,U}$, the output of CSTA can be calculated as $\bm{v}_{\rm f}^{\rm out,C}$ and $\bm{s}_{\rm [CLS]}^{\rm out,C}$, respectively. Thus, the calculations of the whole CSTA can be summarized as:
\begin{multline}
(\bm{v}_{\rm [CLS]}^{\rm out,C}, \bm{v}_{\rm f}^{\rm out,C}, \bm{s}_{\rm [CLS]}^{\rm out,C}, \bm{s}_{\rm f}^{\rm out,C}) = \\
{\rm CSTA}(\bm{v}_{{\rm [CLS]},L_v}^{\rm out,U},\bm{v}_{{\rm f},L_v}^{\rm out,U},\bm{s}_{{\rm [CLS]},L_s}^{\rm out,U},\bm{s}_{{\rm f},L_s}^{\rm out,U}).
\end{multline}

\subsection{Hierarchical Learning}
We refer to a single-layer structure combining USTA and CSTA as Space-Time Attention (STA). In our \textbf{hierarchical} structure, we define each Hierarchical Space-Time Attention (HSTA) module as containing multiple USTA modules and one CSTA module:
\begin{equation}
{\rm HSTA}_m = [({\rm USTA}_{L_v}, {\rm USTA}_{L_s});{\rm CSTA}]
\end{equation}
where $m^{\rm th}$ is one calculation layer of HSTA. Thus, the outputs set ($\bm{z}_M^{\rm H}=[\bm{v}_{{\rm [CLS]},M}^{\rm out,H}, \bm{v}_{{\rm f},M}^{\rm out,H}, \bm{s}_{{\rm [CLS]},M}^{\rm out,H}, \bm{s}_{{\rm f},M}^{\rm out,H}]$) of the whole operations of our $M$ layer HSTA are summarized as:
\begin{equation}
\bm{z}_M^{\rm H} =
\begin{cases} 
\bm{v}_{\rm [CLS]}^{\rm out,C}, \bm{v}_{\rm f}^{\rm out,C}, \bm{s}_{\rm [CLS]}^{\rm out,C}, \bm{s}_{\rm f}^{\rm out,C}, &M=1, \\
{\rm HSTA}_{M}(\bm{z}_{M-1}^{\rm H}), &M>1.
\end{cases}
\label{eq:hsta}
\end{equation}

For training, we select the ``[CLS]s'' of two different modalities to predict the MEs. Specifically, given the outputs of the last HSTA of video and special data as $\bm{v}_{{\rm [CLS]},M}^{\rm out,H}$ and $\bm{s}_{{\rm [CLS]},M}^{\rm out,H}$, respectively, we first concatenate them and use \textbf{Mean Squared Error} (MSE) loss to measure distance between the prediction and the label $Y$:
\begin{equation}
\mathcal{L} ={\rm MSE}(P_\theta([\bm{v}_{{\rm [CLS]},M}^{\rm out,H} || \bm{s}_{{\rm [CLS]},M}^{\rm out,H}]),Y),
\label{eq:y_pred}
\end{equation}
where $P_\theta$ is the prediction function with parameter $\theta$.
\section{Experiments}
In this section, we conduct experiments to evaluate the effectiveness of our proposed method. First, we introduce the experimental settings. Then analyze the effects of different modules of our method. Finally, we compare other state-of-the-art methods with ours. 
Our experiments are intended to address the following research questions (\textbf{RQ}s):

\noindent\textbf{RQ1:} What are the effects of unimodal and crossmodal space-time attentions?

\noindent\textbf{RQ2:} How does hierarchical learning influence the micro-expression recognition results?

\noindent\textbf{RQ3:} How does the performance comparison between our method and the state-of-the-art methods?

\subsection{Experimental Settings}
\subsubsection{Dataset}
We evaluate our method on four benchmark datasets. Specifically, \textbf{CASME3} \cite{li2022cas} is the largest MER dataset. It includes about 1,000 manually annotated MEs with seven expressions (``Happiness'', ``Anger'', ``Sad'', ``Surprise'', ``Fear'', ``Disgust'' and ``Others''). \textbf{CASME II} \cite{yan2014casme}, \textbf{SMIC} \cite{li2013spontaneous}, and \textbf{SAMM} \cite{davison2016samm} contain 247, 161, and 159 MEs videos, respectively. For the CASME II, SMIC, and SAMM datasets, we follow the experimental settings in the MEGC2019 Challenge \cite{see2019megc} to map these datasets into three general categories: ``Negative'', ``Positive'', and ``Surprise''. More details of experimental datasets can be found in our cited work.

\subsubsection{Evaluation Metrics}
Following the metrics in MER2019 challenge \cite{see2019megc}, we use three common evaluation metrics \textbf{unweighted F1-scores} (UF1), \textbf{unweighted average recall} (UAR), and extra \textbf{accuracy} (ACC) to measure the effectiveness of our method. These metrics are calculated as:
\begin{equation}
\begin{aligned}
{\rm UF1} &= {\textstyle \sum_{i=1}^{C}}((2 {\rm TP}_i)/(2 {\rm TP}_i + {\rm FP}_i+{\rm FN}_i))/C, \\
{\rm UAR} &= {\textstyle \sum_{i=1}^{C}}({\rm TP}_i/{\rm N}_i)/C, \\ 
{\rm ACC} &= \textstyle (\sum_{i=1}^{C} {\rm TP}_i)/( \sum_{i=1}^{C}{\rm N}_i),
\end{aligned}
\end{equation}
where $C$ is the total number of the micro-expression categories, ${\rm TP}_i$, ${\rm FP}_i$, and ${\rm FN}_i$ is the number of true positive, false positive, and false negative samples of the $i^{\rm th}$ category, respectively, ${\rm N}_i$ is the sample number of the $i^{\rm th}$ category.

During the testing phase, we employ the \textbf{leave-one-subject-out cross-validation} strategy (LOSO) \cite{li2022mmnet} to compare with other methods. Specifically, for each iteration, one of the subsets is randomly selected to be used as the test set, and the remaining subsets are used as the training set. 
However, this strategy is time-consuming. Therefore, we use the more efficient and less resource-intensive \textbf{$K$-fold cross-validation} strategy \cite{zhao2023dfme} in our ablation studies to help us find the appropriate parameters. 

\subsubsection{Implementation Details}
Due to the significant variation in sample numbers across different categories in MEs datasets, we employ balanced sampling during training to ensure uniform exposure of the model to approximately the same number of samples from each category. In our experiments, we use a batch size of 32, conduct 150 training epochs, set the base learning rate is 5e-5, and the weight decay is 0.05. To better train the parameters, we employ a warm-up learning strategy in the first 5 epochs. Specifically, we set the initial learning rate 1e-6 and gradually increase until it reaches the basic learning rate. 




\subsection{Ablation Studies}
In the ablation study, we use CASME3 in seven-category classification to evaluate the effectiveness of the different components of HSTA. To minimize computational costs, all experiments in this study are conducted using the \textbf{5-fold cross-validation} strategy. Within each fold, we randomly designate 20\% of the MEs as the test set, while the remaining 80\% constitute the training set, each MEs video appears in the test set once and only once.

\subsubsection{The Effectiveness of USTA (RQ1.a)}

\begin{figure}[t]
    \centering
    \includegraphics[width=0.5\textwidth]{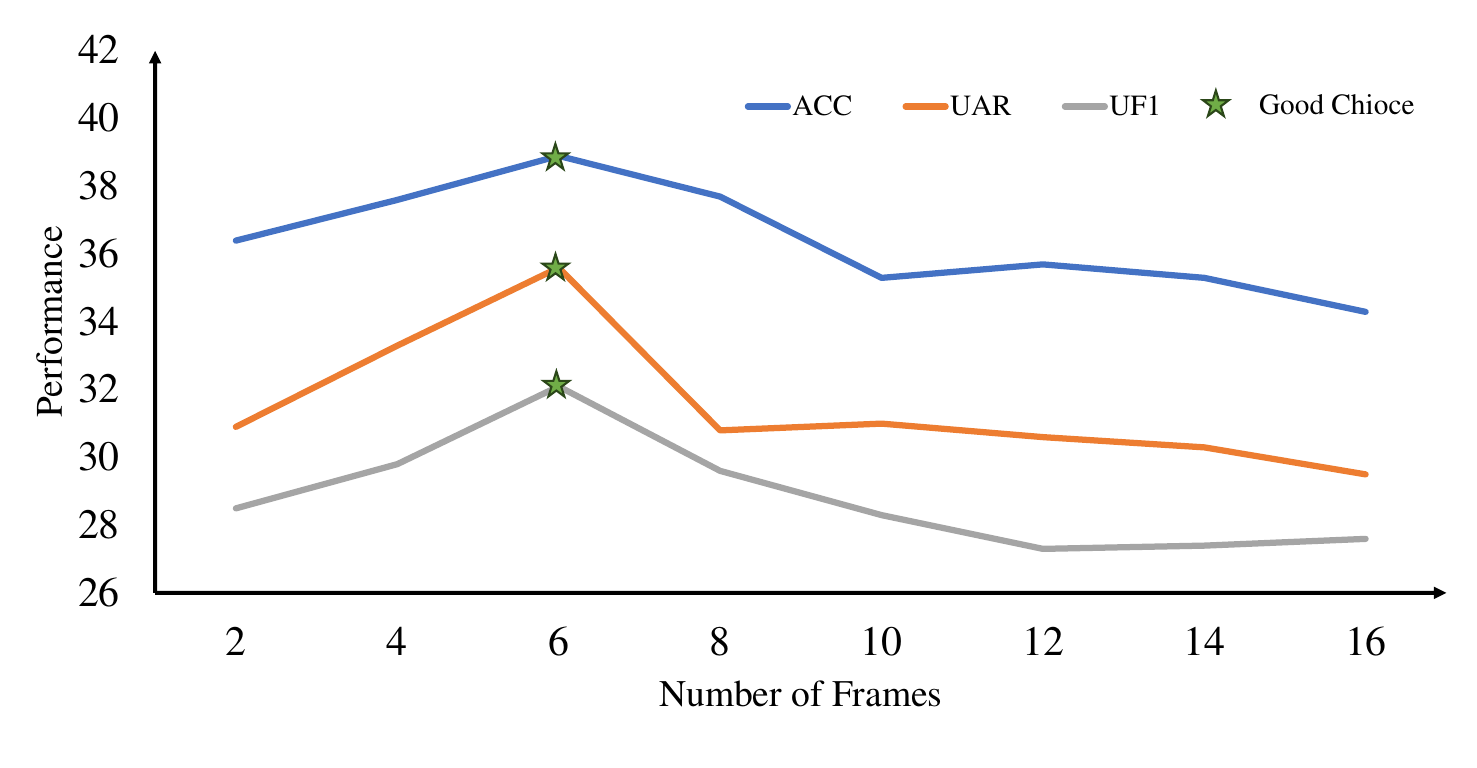}
    \caption{Performances of frames with different numbers in USTA.}
    \label{fig:exp2}
\end{figure}

\begin{table}
    \centering
\begin{tabular}{c|ccc|ccc} 
\hline\hline
\multirow{2}{*}{\#$L$} & \multicolumn{3}{c|}{$L_v$}                    & \multicolumn{3}{c}{$L_s$}                      \\ 
\cline{2-7}

                   & ACC           & UAR           & UF1                     & ACC           & UAR           & UF1                   \\ 
\hline
0                  & 15.8          & 14.3          & 16.1          & 15.8          & 14.3          & 16.1          \\
1                  & 37.5          & 32.4          & 31.1          & 37.9          & 32.1          & 29.4           \\
\hline
2                  & 36.5          & 32.6          & 29.8          & 38.8          & 32.6          & 30.8           \\
4                  & 38.0          & 33.1          & 29.8          & 39.3          & 32.9          & 31.0           \\
6                  & 41.6          & 33.6          & 32.0          & 39.2          & 33.9          & 31.0           \\
\textbf{8}         & \textbf{42.1} & \textbf{36.7} & \textbf{33.5} & 39.5 & \textbf{35.5} & \textbf{32.2}  \\
10                 & 40.0          & 35.5          & 32.0          & \textbf{40.9 }         & 34.1          & 31.8           \\
12                 & 40.1          & 35.6          & 31.8          & 39.5          & 34.0          & 31.4           \\
\hline\hline
    \end{tabular}
    \caption{Performance of different $L_s$ and $L_v$. }
    \label{tab:p_vs_lV_ls}
\end{table}

We conduct experiments solely using USTA. For efficiency and to expedite subsequent computational processes, we initially employ a single-layer USTA to determine the optimal number of uniformly sampled frames for video frames set $\mathcal{D}_{v}$. The results are presented in Figure \ref{fig:exp2}. As illustrated in the figure, varying the frame count significantly impacts the experimental outcomes. Performance initially improves with an increase in frame count but subsequently diminishes. An excessive number of frames not only increases computational overhead but also reduces performance. This can be attributed to the introduction of redundant information by too many frames, which may skew the capture of finer details. Therefore, in our subsequent experiments, we validate our model using \textbf{six} frames.

Next, we verify the impact with and without USTA. We use the Apex frame and Onset frame as special frames in our experiments. Without USTA means we set the total cascaded USTA layer count $L_v=0$ for video frames set $\mathcal{D}_{v}$ in Eq.~\eqref{eq:lvls} and $L_s=0$ for special frames set $\mathcal{D}_{s}$. We also investigate the relationship of different USTA layer counts by setting the number of layers from 1 to 12. The specific results are shown in Table \ref{tab:p_vs_lV_ls}, \#$L$ is the number of $L_v$ or $L_s$. When no USTA layers are included, the performance is extremely poor. However, introducing just a single-layer USTA led to a significant performance improvement (ACC from 16\% $\to$ 38\%). The improvement observed after the inclusion of USTA indicates the critical importance of temporal modeling in MER. It also demonstrates the effectiveness of our USTA.
As $L_v$ or $L_s$ increase, performance improves continuously. However, when $L_v$ or $L_s$ become too large, It leads to overfitting, resulting in an initial increase followed by a decrease.


\subsubsection{The Effectiveness of CSTA (RQ1.b)}
We utilize only CSTA structure and skip the USTA to verify our CSTA. Then We use a structure where both $L_v$ and $L_s$ are 1 to compare with the structure USTA coordinates with CSTA.
In Table \ref{tab:Effectiveness of CSTA and coordination}, the comparison of settings a1 and a2 reveals that employing only the CSTA structure yields a certain level of performance compared to without CSTA. The single application of CSTA alone does not surpass using USTA alone as observed in settings a3 and a4. Whether it is USTA or CSTA, single use results in unsatisfactory performance. Furthermore, the setting a5, which utilizes both video frames and special frames without a robust fusion mechanism, leads to a decline in performance.  In contrast, the combination of CSTA with USTA in setting a6, leads to a significant improvement. This not only demonstrates the effectiveness of CSTA but also indicates that CSTA and USTA work well together. We believe that only after USTA establishes connections between subtle facial movements and specific facial areas, CSTA can achieve higher-quality fusion. Here different modalities also learn aspects of content from each other, enriching their ``[CLS]'' token's content.

\begin{table}[t]
    \centering
    \begin{tabular}{c|ccc|rrr}
        \hline\hline
        Setting &CSTA& $L_v$&$L_s$ & ACC & UAR & UF1 \\
        \hline
        a1&\text{\texttimes}&0&0    & 15.8 & 14.3 & 16.1 \\        
        a2&\text{\checkmark}&0&0    & 35.5 & 30.9 & 28.2 \\
        a3&\text{\texttimes}&1&0    & 37.5 & 32.4 & 31.1 \\
        a4&\text{\texttimes}&0&1    & 37.9 & 32.5 & 29.4 \\
        a5&\text{\texttimes}&1&1    & 37.1 & 32.1 & 29.7 \\
       \textbf{a6}& \textbf{\text{\checkmark}} & \textbf{1} & \textbf{1} & \textbf{40.2} & \textbf{33.9} & \textbf{31.6} \\

        \hline\hline
    \end{tabular}
    \caption{Performances of CSTA and its coordination.}
    \label{tab:Effectiveness of CSTA and coordination}
\end{table}

\begin{table}[t]
    \centering
    \begin{tabular}{c|c|@{\hskip 1pt}c@{\hskip 1pt}cc|ccc}
        \hline\hline
       Setting& HSTA & &$L_{s}$ & $L_{v}$ & ACC & UAR & UF1 \\
        \hline
        b1&1 & &1 & 1 & 40.2 & 33.9 & 31.6 \\
        b2&2 & $\blacksquare$&\textcolor{blue}{1} & \textcolor{blue}{1} & 40.8 & 34.1 & 32.2 \\
       b3 &3 & $\blacksquare$&\textcolor{blue}{1} & \textcolor{blue}{1} & 41.3 & 34.9 & 31.3 \\
        \hline
       b4 & 3 &$\bullet $&\textcolor{blue}{1} &\textcolor{blue}{2} & 40.7 & 35.1 & 32.0 \\
        b5&3 & &1 & 3 & 41.0 & 36.1 & 33.0 \\
        b6&3 & &1 & 4 & 40.3 & 36.0 & 33.2 \\
        b7&3 & &1 & 5 & 40.7 & 35.8 & 32.6 \\
        b8&3 & &2 & 1 & 39.8 & 34.3 & 31.6 \\
       b9 &3 & $\blacktriangle$&\textcolor{blue}{2} & \textcolor{blue}{2} & \textbf{42.4} & \textbf{36.6} & \textbf{34.8} \\
       b10 &3 && 3 & 1 & 41.2 & 34.5 & 32.2 \\
        b11&3 & &3 & 2 & 41.0 & 35.3 & 30.8 \\
        b12&3 & &3 & 3 & 40.8 & 35.2 & 32.8 \\
        \hline\hline
    \end{tabular}
    \caption{Performances of HSTA and its coordination.}
    \label{tab:exp3}
\end{table}

\subsubsection{The Effectiveness of Hierarchical Learning (RQ2)}
Here we evaluate the performance of hierarchical HSTA, as detailed in Table \ref{tab:exp3}. Comparisons with different HSTA layer settings b1, b2, and b3 in Table \ref{tab:exp3} reveal that HSTA methods outperform the single-layer STA unit. The performance of the USTA is influenced by $L_v$ and $L_s$. Maintaining $L_s$ constant while increasing $L_v$ initially enhances performance then decreases (as shown from b3 to b7). In contrast, keeping $L_v$ constant while increasing $L_s$ leads to a continuous decrease in performance. Our experiments suggest that performance is better when $L_v > L_s$ compared to $L_v < L_s$. $L_v=L_s$ also provides satisfactory results, likely due to the larger volume of information from video frames matched by greater computational capacity when $L_v \geq L_s$. Then we explore the relationship between the HSTA layers value $M$ and performance using three combinations of $L_v$ and $L_s$: $\langle \blacksquare L_v=1, L_s=1 \rangle$,  $\langle \bullet L_v=2 ,L_s=1 \rangle$ and  $\langle \blacktriangle L_v=2,L_s=2 \rangle$ highlighted in \textcolor{blue}{blue} in Table \ref{tab:exp3}. Subsequently, we determine the optimal HSTA layers $M$ for these combinations, with results presented in Table \ref{tab:Performances with different numbers of HSTA.}. An interesting observation is that our method performs best on the CASME3 dataset when the total value of $L_v$ or $L_s$ approximates 8, i.e., when $M \times L \approx 8$ performs best. Here $L$ is the number of $L_v$ or $L_s$. For instance, configurations such as $\langle M = 8, L = 1 \rangle$ and $\langle M = 4, L = 2 \rangle$ exhibit good performance, without requiring both $L_v$ and $L_s$ to be set at 8. Considering that $L_v$ corresponds to the processing of larger video frames as mentioned above, we prefer configurations where $L_v > L_s$. Balancing computational load and performance, a configuration like $\langle L_v=2, L_s=1, M=4\rangle$, highlighted in \textcolor{blue}{blue} in Table \ref{tab:Performances with different numbers of HSTA.}, achieves an effective balance between accuracy and computational effort. We plan to apply these optimally determined parameter values in our subsequent experiments.


\begin{table}[t]
    \centering
    \begin{tabular}{c@{\hskip 1pt}|c@{\hskip 5pt}c@{\hskip 5pt}c|c@{\hskip 5pt}c@{\hskip 5pt}c|c@{\hskip 5pt}c@{\hskip 5pt}c}
        \hline\hline
       \multirow{2}{*}{$M$} & \multicolumn{3}{c|}{$\blacksquare L_v$=1 $L_s$=1}      & \multicolumn{3}{c|}{$\bullet L_v$=2 $L_s$=1}& \multicolumn{3}{c}{$\blacktriangle L_v$=2 $L_s$=2}                                    \\ 
\cline{2-10}
                   & ACC           & UAR           & UF1                     & ACC           & UAR           & UF1    & ACC           & UAR           & UF1                \\ 
        \hline
        
        1 & 40.2 & 33.9 & 31.6 & 41.0   & 35.7 & 31.2 & 40.7 & 33.4 & 31.1 \\
        
        2 & 40.8 & 34.1 & 32.2 & 42.2 & 36.4 & 33.5 & 40.0   & 34.5 & 31.2 \\
        3 & 41.3 & 34.9 & 31.3 & 40.7 & 35.1 & 32.0   & 42.4 & 36.6 & \textbf{34.8} \\
        4 & 40.5 & 34.8 & 31.9 & \textcolor{blue}{42.6} & \textbf{\textcolor{blue}{38.0 }} & \textbf{\textcolor{blue}{35.1}} & 42.2 & 36.9 & 33.6 \\
        5 & 41.2 & 36.4 & 33.6 & 40.9 & 36.0   & 32.5 & \textbf{43.1} & \textbf{37.4} & 33.6 \\
        6 & 40.7 & 36.4 & 32.9 & 41.9 & 36.2 & 32.9 & 42.1 & 35.9 & 33.0 \\
        7 & 41.1 & 36.0 & 33.1 & \textbf{42.8} & 37.8 & 35.1 & 41.1 & 36.1 & 32.4 \\
        8 & \textbf{42.8} & \textbf{38.6} & \textbf{35.1} & 41.0 & 36.0   & 33.8 & 40.6 & 36.3 & 32.6 \\
        10 &42.6 & 35.7 & 33.7 & 41.8 & 36.8 & 33.6 & 41.5 & 36.4 & 33.0 \\
12 & 42.4 & 38.5 & 35.1 & 41.3 & 36.1 & 33.4 & 42.2 & 37.0 & 33.3 \\
        \hline\hline
    \end{tabular}
    \caption{Performances with different numbers of HSTA.}
    \label{tab:Performances with different numbers of HSTA.}
\end{table}

\subsection{Comparisons with Other Methods (RQ3)}
We compare our approach with traditional hand-crafted methods, mainstream approaches in recent years and some latest comprehensive methods on the classic MER datasets detailed in three-category classification in Table \ref{tab:Performance comparison of the SOTA methods}. It is observed that our model outperforms any hand-crafted method like LBP-TOP \shortcite{yan2014casme}, Bi-WOOF \shortcite{liong2018less} and those that rely solely on special frames as CapsuleNet \shortcite{van2019capsulenet}, MMNet \shortcite{li2022mmnet} or optical flow Dual-Incep \shortcite{zhou2019dual}, Dual-ATME \shortcite{zhou2023dual}. Compared to other models utilizing temporal information STSTNet \shortcite{liong2019shallow}, MERSiamC3D \shortcite{zhao2021two}, our model also demonstrates higher performance across all datasets. Our approach also outperforms some of the latest comprehensive methods, such as FRL-DGT \shortcite{zhai2023feature}. When compared with Micro-BERT\cite{nguyen2023micron}, there are instances, such as in smaller-scale datasets like SMIC, our model's performance is not as high. However, our computing costs are only about 1/96 of Micro-BERT's. We speculate poor performance here is due to the small data size and the consequential larger impact of randomness. Therefore, we conduct further tests on the larger and more diverse CASME3 dataset, as shown in Table \ref{tab:MER on the CASME3}. Compared to the benchmark method RGB-D\shortcite{li2022cas} used as our baseline for CASME3, which neither considers temporal information nor incorporates the fusion of crossmodal data, our model exhibits 16.4\% performance improvement (UF1 17.7\% $\to$ 34.1\%). So when the data size is sufficiently large, our model's performance significantly surpasses other methods, without the need for the extensive and resource-intensive pre-training required by Micro-BERT.

\begin{table}[t]
\centering

\begin{tabularx}{\columnwidth}{@{\hskip 1pt}l@{\hskip 1pt}|c@{\hskip 1pt}c|c@{\hskip 1pt}c|c@{\hskip 1pt}c@{\hskip 0pt}}
\hline\hline
       \multirow{2}{*}{Method} & \multicolumn{2}{@{\hskip 0pt}c@{\hskip 0pt}|}{SMIC}      & \multicolumn{2}{c|}{SAMM}& \multicolumn{2}{c}{CASMEII} \\

 &UF1&UAR&UF1&UAR&UF1&UAR\\

\hline

LBP-TOP \shortcite{yan2014casme}          &20.0& 52.8& 39.5&41.0 & 70.3 &74.3 \\
Bi-WOOF \shortcite{liong2018less}          & 57.3 &58.3 & 52.1& 51.4 & 78.1& 80.3 \\ 
\hline
CapsuleNet \shortcite{van2019capsulenet}   & 58.2& 58.8 & 62.1& 59.9 & 70.7 &70.2 \\
MMNet \shortcite{li2022mmnet}              & 44.1& 43.8 & 32.6& 34.2 & 71.9 &89.9 \\ 
\hline
Dual-Incep \shortcite{zhou2019dual}        & 57.1 &57.1 & 49.3& 49.6 & 75.4 &77.4 \\
Dual-ATME \shortcite{zhou2023dual}         & 64.6 &65.8 & 56.2& 53.8 & 76.5 &75.1 \\ 
\hline
STSTNet \shortcite{liong2019shallow}       & 68.0 &70.1 & 65.9& 68.1 & 83.8 &86.9 \\
MERSiamC3D \shortcite{zhao2021two}         & 73.6& 76.0 & 74.8 &72.8 & 89.2& 88.7 \\ 
\hline
FRL-DGT \shortcite{zhai2023feature}        & 74.3 &74.9 & 77.2 &75.8 & 91.9 &90.3 \\
Micro-BERT \shortcite{nguyen2023micron}    & \textbf{85.5 }&\textbf{83.8} & 83.9& \textbf{84.8} & 90.3 &89.1 \\
\hline
HSTA(Ours) & 84.7  &78.0& \textbf{84.7 }&83.9 & \textbf{92.5 }&\textbf{92.2} \\
\hline\hline
\end{tabularx}

\caption{Comparisons on SMIC, SAMM, and CASME II.}
\label{tab:Performance comparison of the SOTA methods}
\end{table}

\begin{table}[t]
\centering

\begin{tabular}{ll|c|cc} 
\hline\hline
\multicolumn{1}{c}{Method}  &  & Classes & UF1 & UAR   \\ 
\hline
                         
                         FR \shortcite{zhou2022feature}    &  & 3       &  34.9                        &      34.1 \\
HTNet \shortcite{wang2023htnet}    &  & 3       &  57.7                        &      54.2 \\
Micro-BERT \shortcite{nguyen2023micron}          &  & 3       & 56.0                     & 61.3  \\
HSTA(Ours) &  & 3       & \smash{\textbf{ 59.3 }}                       &   \textbf{ 61.8 }  \\ 
\hline

                             RGB \shortcite{li2022cas}           &  & 7       & 17.6                     & 18.0  \\
                             RGB-D \shortcite{li2022cas}           &  & 7       & 17.7                     & 18.3  \\
Micro-BERT \shortcite{nguyen2023micron}           &  & 7       & 32.6
& 32.5  \\
HSTA(Ours) &  & 7       &  \textbf{34.1}                      &  \textbf{ 35.8 }  \\
\hline\hline

\end{tabular}
\caption{Comparisons with others on CASME3.}
\label{tab:MER on the CASME3}
\end{table}

\subsection{Additional Exploration}

We discover that benchmark datasets contain a wealth of additional data, such as macro-expressions (MaE) and objective classes (OC) which utilize objective facial muscle motion blocks - action units (AUs) - as proposed by \cite{davison2018objective} to categorize MEs rather than relying on annotator or subject's subjective judgment in CASME3. However, the use and exploration of these data have been limited in previous studies, which we aim to investigate their effectiveness. Recent MER methods tend to favor the utilization of optical flow (OF), but it is not essential. Due to the high flexibility of our model's input, we replace the input of the special frames (SF) with optical flow to compare their effectiveness. Based on these considerations mentioned above, we conduct the following experiments: firstly, we use SF as the input as our baseline and then replace them with the OF; secondly, we employ a more rational label categorization method called objective classes to divide MEs categories, avoiding subjective judgment; thirdly, we explore the effect of incorporating MaE as additional data for training. The performance of these three approaches is presented in Table \ref{tab:Addition Studies}.

Using optical flow data extracted from special frames leads to a decrease in performance compared to directly using special frames in Table \ref{tab:Addition Studies}. When we utilize more objective classes based on AUs for a seven-category classification, there is a significant performance improvement (UF1 35.1\% $\to$ 43.0\%). This further demonstrates the precision of our model in capturing subtle, objective facial movements. Given the limited amount of ME data, the model does not fully realize its potential, particularly when encountering unfamiliar facial types or expressions. However, by utilizing larger data from the CASME3 dataset for auxiliary training, we overcome the limitations (UF1 35.1\% $\to$ 52.7\%). Our model excels with both subjective labels and objective classes. In summary,  we can attain even higher performance by the often-neglected data in datasets which are also of great value, and our method is a universally applicable one.

\begin{table}
\centering

\begin{tabular}{c@{\hskip 4pt}c@{\hskip 4pt}c@{\hskip 4pt}c@{\hskip 5pt}|c@{\hskip 5pt}|rrr}

\hline\hline
SF &OF & MaE & OC &Classes& ACC  & UAR  & UF1  \\
\hline
\text{\checkmark}  &\text{\texttimes}            & \text{\texttimes}        &  \text{\texttimes}                 &7& 42.6 & 38.0 & 35.1 \\
 \text{\texttimes}  &\text{\checkmark}            & \text{\texttimes}          &  \text{\texttimes}                &7 & 40.1 & 36.4 & 35.4 \\
 \text{\texttimes}  &\text{\texttimes}            & \text{\texttimes}          &   \text{\checkmark}               &7 & 52.3 & 43.2 & 43.0 \\
  \text{\texttimes}  &\text{\checkmark}             & \text{\checkmark}           & \text{\checkmark}                  &7 & 48.6 & 39.3 & 41.8 \\
\textbf{\text{\checkmark}} & \textbf{\text{\texttimes}} & \textbf{\text{\checkmark}} & \textbf{\text{\checkmark}} & 7 & \textbf{69.8} & \textbf{51.7} & \textbf{52.7} \\

\hline\hline
\end{tabular}
\caption{Additional exploration on other data in CASME3 dataset.}
\label{tab:Addition Studies}
\end{table}


\section{Conclusion}
In this paper, we propose a hierarchical attention strategy for different modalities to tackle the MER problem. Specifically, (1) Unimodal space-time attention (USTA) is used to capture temporal information in MER. (2) Crossmodal space-time attention (CSTA) is designed to fuse the different modalities while maintaining their uniqueness. (3) The hierarchical structure based on USTA and CSTA is proposed to grasp deeper facial cues. The extensive experiments have demonstrated the effectiveness of our proposed method. In addition, we verify the generalizability of our method on different types of additional data contained in a benchmark dataset. 

Note that our method relies on a cascaded structure of USTA with CSTA, which may increase the computational complexity. In future work, we will focus on more efficient space-time attention methods to accelerate the recognition process of MEs.

  
\clearpage
\bibliographystyle{named}
\bibliography{ijcai.bib}

\end{document}